\documentclass[runningheads]{llncs}
\usepackage{graphicx}
\usepackage{algorithm}
\usepackage{algorithmic}
\usepackage{graphicx}
\graphicspath{}
\usepackage{capt-of}
\usepackage{float} 
\usepackage{subfigure} 
\usepackage{url}
\usepackage{booktabs} 
\begin{document}
\title{Trajectory planning in Dynamics Environment: Application for Haptic Perception in Safe Human-Robot Interaction}

\titlerunning{Trajectory planning in Dynamics Environment}

\author{A. Gutierrez, V.K Guda, S. Mugisha, C. Chevallereau, D. Chablat}
\authorrunning{A. Gutierrez, V.K Guda, S. Mugisha, C. Chevallereau, D. Chablat}
\institute{Laboratoire des Sciences du Num\'erique de Nantes, UMR CNRS 6004, Nantes 44300, France \\
{\it damien.chablat@cnrs.fr} }
\maketitle  
\begin{abstract}
In a human-robot interaction system, the most important thing to consider is the safety of the user. This must be guaranteed in order to implement a reliable system. The main objective of this paper is to generate a safe motion scheme that takes into account the obstacles present in a virtual reality (VR) environment.
The work is developed using the MoveIt software in ROS to control an industrial robot UR5. Thanks to this, we will be able to set up the planning group, which is realised by the UR5 robot with a 6-sided prop and the base of the manipulator, in order to plan feasible trajectories that it will be able to execute in the environment.
The latter is based on the interior of a vehicle, containing a user (which would be the user in this case) for which the configuration will also be made to be taken into account in the system.
To do this, we first investigated the software's capabilities and options for path planning, as well as the different ways to execute the movements. We also compared the different trajectory planning algorithms that the software is capable of using in order to determine which one is best suited for the task. Finally, we proposed different mobility schemes to be executed by the robot depending on the situation it is facing. The first one is used when the robot has to plan trajectories in a safe space, where the only obstacle to avoid is the user's workspace. The second one is used when the robot has to interact with the user, where a mannequin model represents the user's position as a function of time, which is the one to be avoided. 
\end{abstract}
%%%%%%%%%%%%%%%%%%%%%%%%%%%%%%%%%%%%%%%%%%%%%%%%%%%%%%%%%%%
\section{Introduction}
%%%%%%%%%%%%%%%%%%%%%%%%%%%%%%%%%%%%%%%%%%%%%%%%%%%%%%%%%%%
In human-robot interaction systems, knowing how to compute a path for the robot to follow, while taking into account the human position, is a crucial task to ensure the safety of the individuals around the robot. This is where path and trajectory planning plays its role in the field of robotics, where achieving real-time behaviour is one of the most challenging problems to solve. The result is a constant demand for research into more complex and efficient algorithms that allow robots to perform tasks at higher speeds, reducing the time they need to complete them, resulting in increased efficiency. But this also comes at a cost: to achieve higher speeds and shorter times, robot actuators must work under more demanding conditions that can shorten their overall life or even damage their structure. High operating speeds can also affect the accuracy and repeatability of manipulators. Therefore, it is important to generate well-defined trajectories that can be executed at high speeds without generating high accelerations (to avoid robot wear or end effector vibrations during stopping).
Path planning is the generation of a geometrical path from an initial point to an end point and the calculation of the crossing points between them. Each point of the generated trajectory is supposed to be reached by the robot end effector through a specific movement. When the robot is supposed to interact with a human, its velocity and acceleration must be zero at the end of the trajectory.
Another important element to take into account is the environment in which the task or the movement is going to be performed. This is what allows the system to identify the robot's environment and the colliding objects that might be present, thus determining the areas in which the robot must be constrained or limited to ensure the safety of the user.

The Lobbybot project is a project that allows interaction between a user and a cobot. These interactions allow for the creation of a touch-sensitive interface or intermitant contact interface (ICI). The scenario used allows the user to be inside a car with the possibility to interact with its environment by getting a sensory feedback of the different surfaces thanks to a 6 faces prop providing the different textures. Due to the immersion of the user via a VR headset, the system must ensure the safety of the user, as he cannot see the location of the robot. Therefore, it is necessary to implement trajectory planning techniques to be able to avoid unwanted interactions between the robot and the user. To do this, the system must take into account the obstacles present (environment or user). A virtual mannequin is modelled using data from the HTC Vive trackers which provide an estimate of the user's position, and will give the system a model to plan the movements. Thus, the goal of the LobbyBot project is to provide an immersive VR system that is safe for the user and gives them the ability to interact with the environment at different locations, providing a new level of interaction between VR environments and the real world. 
%%%%%%%%%%%%%%%%%%%%%%%%%%%%%%%%%%%%%%%%%%%%%%%%%%%%%%%%%%%
\section{State of the art}
%%%%%%%%%%%%%%%%%%%%%%%%%%%%%%%%%%%%%%%%%%%%%%%%%%%%%%%%%%%
\subsection{Intermittent contact interface}
%%%%%%%%%%%%%%%%%%%%%%%%%%
In the area of human-robot interaction and haptic perception, the ability to reproduce the sense of touch to appreciate different textures and motion sensations through the use of cobots has been addressed in \cite{ENTROPiA}, where a rotatable metaphorical accessory approach (ENTROPiA) has been proposed to provide an infinite surface haptic display, capable of providing different textures to render multiple infinite surfaces in VR (virtual reality). Studies in \cite{AlteredSPF} \cite{EnhancedSPF} have focused on the perception of stiffness, friction, and shape of tangible objects in VR using a wearable 2-DoF (degrees of freedom) tactical device on a finger to alter the user's sense of touch. In \cite{Guda,Lobbybot}, a 6-DoF cobot is used in a VR environment to simulate the interior of a car, where interaction between the robot and the user is expected just at specific, instantaneous points. This proposal is to use ICIs (Intermittent Contact Interfaces)  \cite{ICI}  to minimise the amount of human-robot interactions to increase safety. In order to use the proposed implementations in this study in a real-time environment that involves human movement, it is important to ensure the safety of both the user and the robot to avoid potential collisions or accidents. This is where it is necessary to implement proper path and trajectory planning, in order to determine a feasible path to the desired goal, while avoiding interaction with the human until said goal is reached, generating a human-robot interaction just at the desired time.
%%%%%%%%%%%%%%%%%%%%%%%%%%%%%%%%%%%%%%%%%%%%%%%%%%%%%
\subsection{Path Planning}
%%%%%%%%%%%%%%%%%%%%%%%%%%
Path planning refers to the calculation or generation of a geometric path, which connects an initial point to an end point, passing through intermediate via-points. These trajectories are intended to be followed by the end effector of a robot in order to execute a desired task or motion. This geometric calculation is based on the kinematic properties of the robot as well as its geometry (included in its workspace). In the simplest case, path planning is performed within static and known environments. However, this problem can also be generated for robotic systems subjected to kinematic constraints in a dynamic and unknown environment.

\begin{figure}
\centering
\includegraphics[width=8.5 cm]{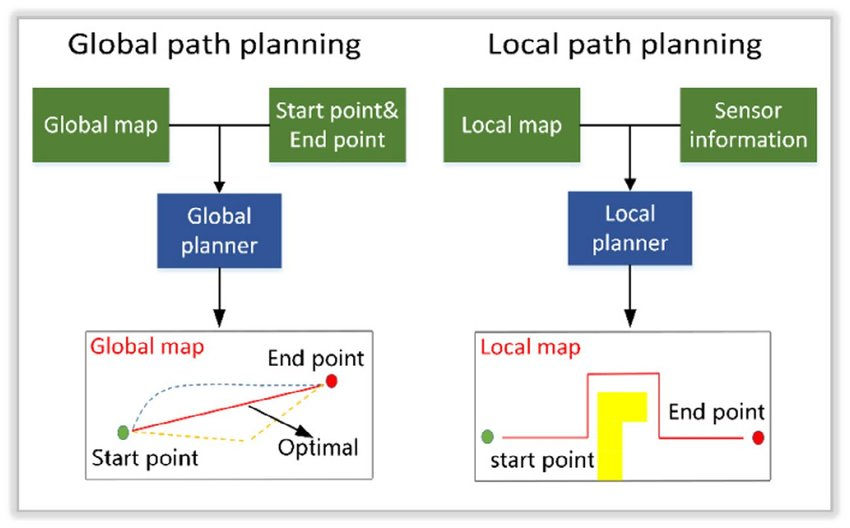}
\caption{Global  \cite{PRM} and local \cite{apf} path planning.}
\label{fig:g_l_pp}
\end{figure}

Path planning can be done using a previously known map. This is called global planning. This method is commonly used to determine the possible paths to follow to reach the final position. It is used in the case of a known and static environment, where the position of the obstacles does not change. This operation can be performed offline, as it is based on previously known information. In the case of dynamic environments, it is necessary to perform local path planning, which relies on sensors or any other type of interface providing data to obtain updated information about the robot's environment. This planning can only be done in real time, as it depends on the dynamic evolution of the environment. Figure \ref{fig:g_l_pp} presents the main differences between local and global path planning \cite{PRM,apf}.

There have been multiple proposals on path planning algorithms over the years. In \cite{PPreview}, one can find a review of the basics and workings of the most common algorithms most commonly found in the robotics literature. The main methods are the following:
\begin{itemize}
    \item The Artificial Potential Fields (APF) approach \cite{apf} introduced by O. Khatib in 1985 and further developed by \cite{apf_volpeThesis} \cite{apf_volpe_paper}. 
    \item The Probabilistic Road-maps approach \cite{PRM} consists in generating random nodes in the configuration space ($C_{space}$) in order to generate a grid (so called, the road-map).
    \item The Cell Decomposition algorithms \cite{CD}.
    \item The Rapidly Exploring Random Trees (\emph{RRTs}) \cite{RRT}, introduced by S. LaValle in 2001 as an optimisation from the classical Random Trees algorithm.
\end{itemize}
%%%%%%%%%%%%%%%%%%%%%%%%%%%%%%%%%%%%%%%
\subsection{Algorithm Comparison}
%%%%%%%%%%%%%%%%%%%%%%%%%%%%%%%%%%%%%%%
In robotics, path planning is one of the most difficult tasks in real-time dynamic environments. Among the presented algorithms, APF and its variations offer a good adaptation to path planning in dynamically changing environments, where any obstacle entering the $C_{space}$ generates a new repulsive field that can be taken into account to generate a new path. But the local minima problem requires the use of alternative algorithms to overcome it.

The case of PRM, it is well known for its ability to find a path without needing to explore the whole $C_{space}$, but it is also a graph based algorithm, which requires the use of shortest path method like \emph{A$^*$}. It works well in static environments and can handle initial and final configuration changes, but if the objects in $C_{space}$ change position, the connections between the nodes must be redone. Some alternatives propose to keep the previously generated nodes and recheck whether they belong to $C_{free}$ or $C_{obs}$, then rebuild the graph based on this information and find a new path. This is also the case for cell decomposition methods, where the graph search has to be reconstructed again. Nevertheless, these methods have proven to be viable options in real time, capable of adapting to a dynamic environment.

Finally, regarding the RRT and RRT* methods and their alternatives, they are known to be good path planning methods, with the limitations that the generated trees are related to the initial configuration and have high computational demands. The proposal of the different alternatives allows to obtain very optimal real-time path planners. The limitations of this type of algorithms are that they require a large memory capacity, as the entire tree must be stored at all times, and that they only work in bounded environments, with unbounded and long distance environments remaining a challenge.
%%%%%%%%%%%%%%%%%%%%%%%%%%%%%%%%%%%%%%%%%%%%%%%%%%%%%%%%%%%
\subsection{Setup of the experimentation}
%%%%%%%%%%%%%%%%%%%%%%%%%%%%%%%%%%%%%%%%%%%%%%%%%%%%%%%%%%%
In this section, we will present the tools used in the development of the project, such as the laboratory system, the software used, a description of the system environment as well as the laboratory setup.
%%%%%%%%%%%%%%%%%%%%%%%%%%%%%%%%%%%%%%%%%%%%%%%%%%%%%%%%%%%
\subsubsection{System Architecture}
%%%%%%%%%%%%%%%%%%%%%%%%%%%%%%%%%%%%%%%%%%%%%%%%%%%%%%%%%%%
The architecture of MoveIt is based on two main nodes, the node \emph{move\_group} and the node \emph{planning\_scene}, which is part of the first one. The \emph{move\_group} node is responsible for obtaining the parameters, configuration and individual components of the robot model being used, in order to provide the user with services and actions to use on the robot.

Within the planners available in the \emph{OMPL} library there are:
\begin{itemize}
    \item PRM methods (PRM \cite{PRM}, PRM* \cite{PRMstar}, LazyPRM \cite{LazyPRM}, LazyPRM* \cite{LazyPRM} \cite{PRMstar}),
    \item RRT methods (RRT \cite{RRT}, RRT* \cite{RRTstar},  TRRT \cite{TRRT}),          BiTRRT \cite{BiTRRT}, LBTRRT \cite{LBTRRT},  RRTConnect \cite{RRTConnect},
    \item Expansive Spacial Trees (EST) methods (EST \cite{EST}, BiEST \cite{EST}).
\end{itemize}
%%%%%%%%%%%%%%%%%%%%%%%%%%%%%%%%%%%%
\subsubsection{Collision detection}
%%%%%%%%%%%%%%%%%%%%%%%%%%%%%%%%%%%%
Collision checking in MoveIt is configured within a planning scene using the CollisionWorld object. Collision checking in MoveIt is performed using the Flexible Collision Library (FCL) package - MoveIt's main collision checking library.
%%%%%%%%%%%%%%%%%%%%%%%%%%%%%%%%%%%%
\subsubsection{Kinematics}
%%%%%%%%%%%%%%%%%%%%%%%%%%%%%%%%%%%%
MoveIt uses a plugin infrastructure, specifically designed to allow users to write their own inverse kinematics algorithms. Direct kinematics and Jacobian search are built into the RobotState class itself. The default inverse kinematics plugin for MoveIt is configured using the KDL numerical solver \cite{KDL} based on Jacobians. This plugin is automatically configured by the MoveIt configuration wizard.
%%%%%%%%%%%%%%%%%%%%%%%%%%%%%%%%%%%%
\subsubsection{ROS-Industrial}
%%%%%%%%%%%%%%%%%%%%%%%%%%%%%%%%%%%%
ROS-Industrial is an open-source project that extends the advanced capabilities of ROS software to industrial hardware and applications. For this project, we used the ROS-Industrial-Universal-Robots metapackage \cite{RI_UR}, which provides and facilitates the main configuration files for the use of Universal Robots cobots in the ROS environment, providing the different descriptions of the robot, configuration files such as joint boundaries, UR kinematics, etc.. This package also facilitates the use of the robot in MoveIt, providing the setup for its use in simulation or in real implementations.
%%%%%%%%%%%%%%%%%%%%%%%%%%%%%%%%%%%%
\subsubsection{HTC Vive}
%%%%%%%%%%%%%%%%%%%%%%%%%%%%%%%%%%%%
The HTC Vive is a motion tracking system that allows users to be immersed in a VR system \cite{HTCVive}. It consists of trackers, which can attach to any rigid object, and work with the VR headset. The tracker creates a wireless connection between the object and the headset and then allows the user to represent the objects movements in a virtual world.
%%%%%%%%%%%%%%%%%%%%%%%%%%%%%%%%%%%%
\subsubsection{Laboratory Setup}
\begin{figure}
    \begin{minipage}[c]{.46\linewidth}
        \centering
        \includegraphics[width=4cm]{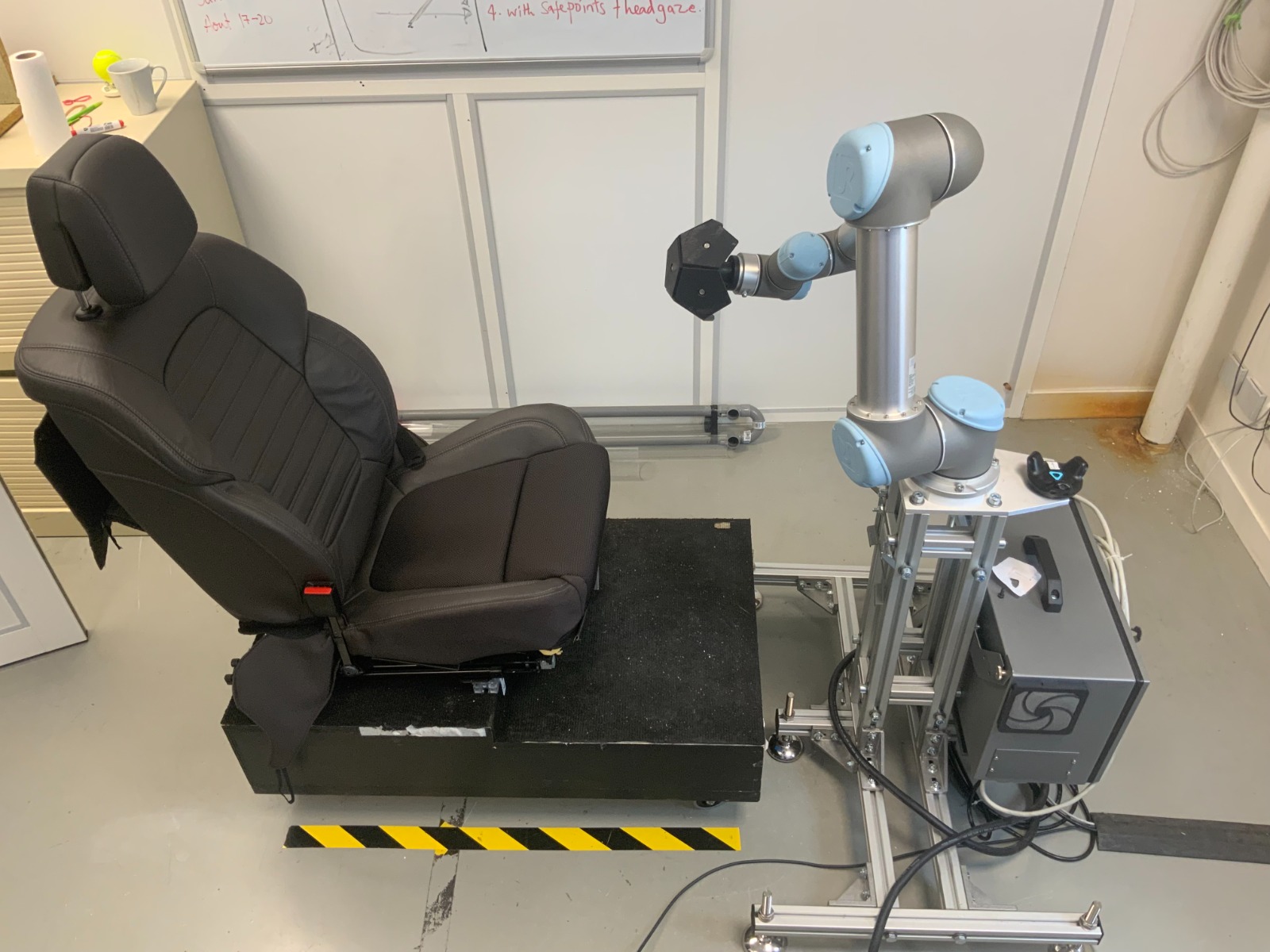}
        \caption{Laboratory Setup}
        \label{fig:lab_su}
    \end{minipage}
    \begin{minipage}[c]{.46\linewidth}
        \centering
         \includegraphics[width=5cm]{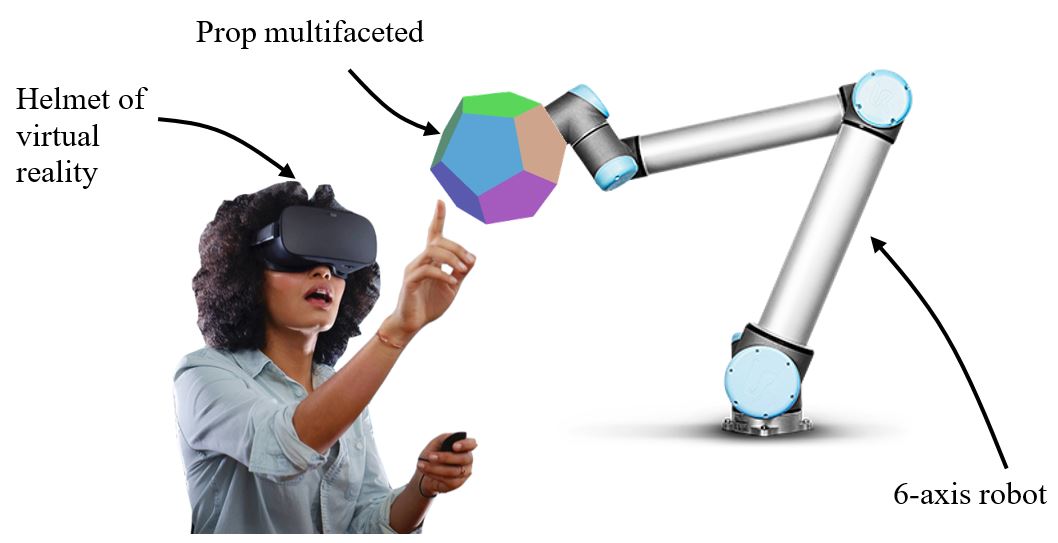}
         \caption{Conceptual scheme of the experimental platform}
         \label{Casque}
    \end{minipage}
\end{figure}

%%%%%%%%%%%%%%%%%%%%%%%%%%%%%%%%%%%%
The laboratory setup consists of a UR5 robotic system and a car chair in a face-to-face configuration (Figure~\ref{fig:lab_su}). The location and height of the robot was determined by \cite{Guda} to be 75 cm above the floor. This position is optimal enough for the robot to reach all the interaction points that the system is interested in reaching. For the user, the VR headset and trackers are attached to the body (the humerus and palms), in order to obtain data and locate the user's location in the VR environment (Figure~\ref{Casque}).
%%%%%%%%%%%%%%%%%%%%%%%%%%%%%%%%%%%%%%%
\section{Selection of the optimal trajectory planning and its application}
%%%%%%%%%%%%%%%%%%%%%%%%%%%%%%%%%%%%%%
We present the setup associated with the choice of the optimal trajectory generator available within the MoveIt software and its application for the LobbyBot project.
%%%%%%%%%%%%%%%%%%%%%%%%%%
\subsection{MoveIt Setup}
%%%%%%%%%%%%%%%%%%%%%%%%%%
The installation of MoveIt consisted of configuring and defining the planning group, as well as making it compatible to work in Gazebo.
The start-up phase was very important to analyse the behaviour of the different movement alternatives found in the MoveIt API. For this, it was important to configure the simulation environment in Gazebo so that we could test without compromising the real robot.
%%%%%%%%%%%%%%%%%%%%%%%%%%
\subsection{Planning group}
%%%%%%%%%%%%%%%%%%%%%%%%%%
The planning\_group is defined as the group of elements  that make up the entire robotic system. These are the UR5 robot, the 6-faced prop and the robot support. These three elements are the ones that the trajectory planning algorithms must consider in order for them to avoid any collision state existing with one of these elements.  The robot support was modelled to match the size of the real system that was optimally defined \cite{Guda}. For the configuration of the plannig\_group, MoveIt has an integrated graphical interface  to create all the configuration files related to the kinematics, controllers, Semantic Robot Description Format (SRDF) and other files for the usage of the robot in ROS. This interface is called \emph{MoveIt Setup Assistant}. The MoveIt Setup Assistant creates all the mentioned files based on the robot description given to it, in this case the UR5 robot description files provided by \cite{RI_UR} where taken and modified to include the robot support  (included in the URDF definition of the robot) and also the mesh file for the prop.
%%%%%%%%%%%%%%%%%%%%%%%%%%
\subsection{User's Model}
\label{sec:person_model}
\begin{figure}
\centering
\includegraphics[width=4.5cm]{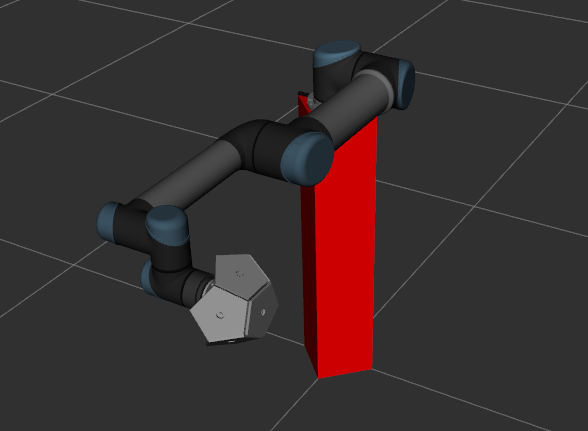}
\includegraphics[width=4.5cm]{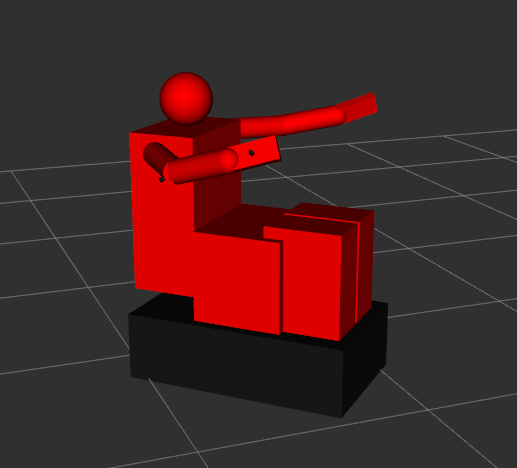}
\caption{Planning group and mannequin model of the user}
\label{fig:mannequin}
\end{figure}

%%%%%%%%%%%%%%%%%%%%%%%%%%
To model the user, a mannequin was defined in a URDF robot model. The main torso of the model is fixed, while the arms are structured as a serial robot with seven revolute joints, where the first three constitute the shoulder, the fourth joint represents the elbow, and the last three revolute joints represent the wrist of the arm. In the model, two small dots have been created in the humerus and palm links, which represent the location of the sensors in the user, as shown in Figure \ref{fig:mannequin}. Regarding the movement of the mannequin model, a kinematic model has been developed in parallel to this project in \cite{mannequinKinematicsVamsi}, where the connection between the sensor data and the model is defined. This will allow the system to recognise the user's movements and represent it in the simulation \ref{fig:mannequin}.
%%%%%%%%%%%%%%%%%%%%%%%%%%
\subsection{Motions}
%%%%%%%%%%%%%%%%%%%%%%%%%%
To generate collision-free trajectories, the different algorithms implemented in the MoveIt API have been analysed. All the tasks related to planning group movement are handled by the move\_group class. By specifying the planning group we want to consider, we are able to use all the different functions that the class offers for it, such as getting information about the current values of the joints, the target, configuring the planning algorithm we intend to use, and performing the planning and execution of the movements in the environment.
%%%%%%%%%%%%%%%%%%%%%%%%%%
\subsubsection{Types of movement}
%%%%%%%%%%%%%%%%%%%%%%%%%%
The move\_group class has the ability to perform path planning through different types of movements. These options can be chosen according to the nature of the task. For example, we can define a given pose in workspace or a desired joint value as the goal. Given the nature of the system, we will work with joint value goals, as we hope to achieve the different points in a specific configuration that provides a higher level of safety to the user (elbow up configuration for the UR5).

Another important feature is the ability to specify whether one wants to achieve each of the requested objectives or not. As the implementation will receive constantly changing goals, the best implementation is to plan and move towards said goal by allowing the system to replan if the goal changes, meaning that we do not need to reach the initial goal. To do this, the move\_group class relies on the $move\_group.execute(my\_plan)$ function to strictly reach the goal and on the $move\_group.asyncExecute(my\_plan)$ function to execute the planned path with the possibility of re-planning during this execution.

\begin{figure}
\centering
\includegraphics[width=5cm]{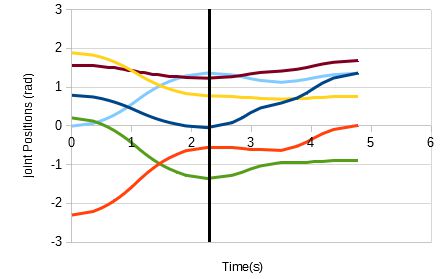}
\includegraphics[width=7cm]{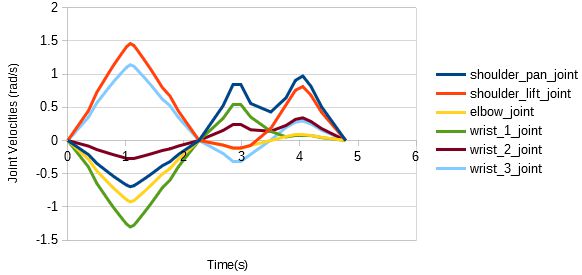}
\caption{Planned paths using $move\_group.execute(my\_plan)$: Back-to-back plans (left) and Velocities for both plans (right).}
\label{fig:2plans_exec}
\end{figure}

In Figure \ref{fig:2plans_exec}, two trajectories are calculated from an initial configuration, to an intermediate goal, and then to a final goal. In this case, by using the function $move\_group.execute(my\_plan)$, we ensure that the robot will completely execute each of the trajectories and achieve both goals. This is illustrated in Figure \ref{fig:2plans_exec}, where the speeds drop to zero as the robot comes to a stop.

\begin{figure}
\centering
\includegraphics[width=5cm]{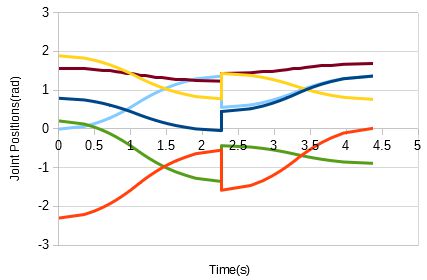} (a)
\includegraphics[width=5cm]{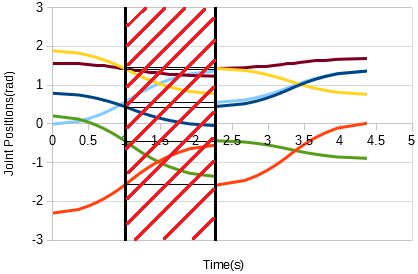}(b)\\
\includegraphics[width=7cm]{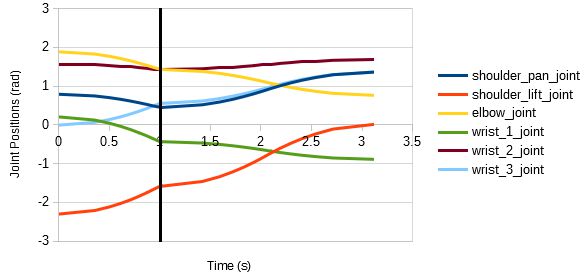}
(c)
\caption{Re-planned paths using $move\_group.asyncExecute(my\_plan)$} (a) Back-to-back plans. (b) Segment not executed due to re-planning. (c) Final executed plan.
\label{fig:2plans_asyncExec}
\end{figure}

In the case of figure \ref{fig:2plans_asyncExec}, we have calculated the same two trajectories as before, but using the function $move\_group.asyncExecute(my\_plane)$, which allows replanning during the execution of the first plane. In this case, in figure \ref{fig:2plans_asyncExec}(a), we can see the two plans one after the other, while in figure~\ref{fig:2plans_asyncExec}(b), we show the representation of the segment that was not executed from the first plan, because a replanning scenario was set up. In this case, the current positions of the first plan were taken as the initial positions for the second plan, resulting in Figure~ \ref{fig:2plans_asyncExec}(c), showing the two plans that were executed.
%%%%%%%%%%%%%%%%%%%%%%%%%%
\subsubsection{Algorithm Selection}
\label{subsec:alg_sel}
\begin{figure}
\centering
\includegraphics[width=7.5cm]{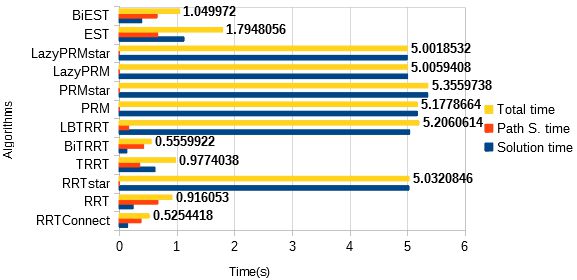}
\caption{Comparison of all planning algorithm's times for nine trajectories}
\label{fig:all_plan_alg}
\end{figure}

%%%%%%%%%%%%%%%%%%%%%%%%%%
Another parameter to select was the planning algorithm that best suited the task. As mentioned earlier, MoveIt has several built-in path planning algorithms that can be used. In order to determine the best option, we went through all the available options and performed a planning task to a desired target configuration, measuring the time required for each algorithm and recording the data. We ran each of the 12 available planning algorithms five times through nine different paths. We then took the average time it took them to find a solution, to simplify the trajectory (only for the algorithms that had this feature) and calculated the total average time. Using this data, we were able to select the algorithms that performed best with the shortest planning times (Figure~\ref{fig:all_plan_alg}).

\begin{figure}
\centering
\includegraphics[width=7.5cm]{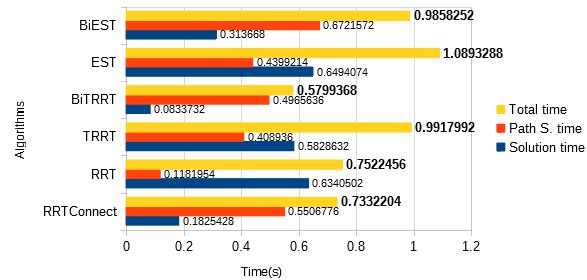}
\caption{Comparison of best planning algorithm's times for 12 trajectories}
\label{fig:all_plan_filt}
\end{figure}

After performing these calculations, given the large difference in planning times for some of the algorithms, we select the six best algorithms to compare them on 12 trajectories (Figure~\ref{fig:all_plan_filt}).

Another analysis that allowed us to select the algorithm which behaved the best for the implementation, was to perform an analysis on the generated trajectories with each one of the algorithms for a fixed task. Based on the six best algorithms from the previous analysis as a starting constraint, we computed the average execution time and via-points number for a set of trajectories. The BiTRRT algorithm wins for both comparisons.

\begin{figure}
    \begin{minipage}[c]{.46\linewidth}
        \centering
        \includegraphics[width=6cm]{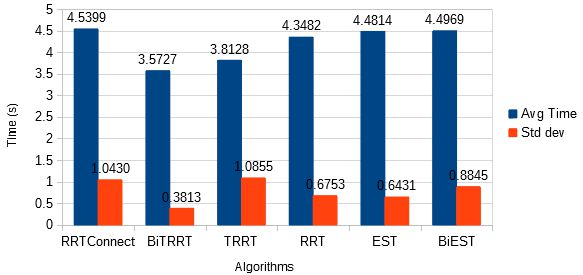}
        \caption{Comparison of average execution times of the algorithms.}
        \label{fig:avg_ex_times}
    \end{minipage}
    \hfill%
    \begin{minipage}[c]{.46\linewidth}
        \centering
        \includegraphics[width=6cm]{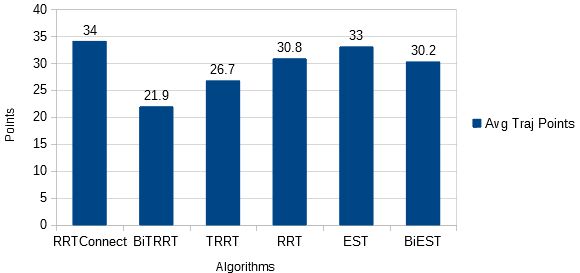}
        \caption{Comparison of average amount of generated via-points.}
        \label{fig:avg_ex_points}
    \end{minipage}
\end{figure}
This analysis was performed for the same trajectories as in the previous graphs for a total of ten iterations for each algorithm, but instead of considering only the computation time (Figure \ref{fig:avg_ex_times}), we also took into account the amount of via-points generated (Figure \ref{fig:avg_ex_points}). Between each via-point, a linear interpolation is performed in the joint space. For theses trajectories, the mannequin was placed in its seat so that it was avoided in the calculation, in order to test each algorithm's ability to plan around it. This also allowed us to see how consistent the behaviour of each algorithm was.
%%%%%%%%%%%%%%%%%%%%%%%%%%
\subsubsection{Unity's Virtual Environment}
%%%%%%%%%%%%%%%%%%%%%%%%%%
In parallel to the development of the project, and to better explain the developed implementation, it is important to specify how it will fit into the project. The system will receive a desired goal configuration which will be the $q_{goal}$ for the planning algorithm, from the current $q_{init}$ configuration. This goal selection is done in Unity by a \emph{Point selection algorithm} which determines the interaction point the user intends to reach \cite{S_V_paper} (Figure~\ref{fig:VR_int_points}).
%%%%%%%%%%%%%%%%%%%%%%%%%%
\subsection{Planning Scene}
%%%%%%%%%%%%%%%%%%%%%%%%%%
For the definition of the planning environment and scene, MoveIt has instances that allow the manipulation and monitoring of the scene to keep it up to date. These instances are : 
\begin{itemize}
    \item PlanningSceneInterface: Is responsible for adding and removing objects in the scene.
    \item PlanningSceneMonitor: Takes care of keeping track of the planning\_scene in order to keep it updated.
\end{itemize}
The last of these instances is absolutely necessary to perform the collision check, as we need to ensure that the scene being processed is the last one available.
%%%%%%%%%%%%%%%%%%%%%%%%%%
\subsection{Mobility Schemes}
%%%%%%%%%%%%%%%%%%%%%%%%%%
Based on the Unity information, two different motion or mobility schemes and scenarios has been proposed depending on the nature of the task we want to achieve at the moment. One for which no interaction with the user is required, and another one for when it is. These two scenarios have their own environment to consider, presenting in general two different behaviours.
%%%%%%%%%%%%%%%%%%%%%%%%%%
\subsubsection{Movement outside user's workspace}
\label{sec:outside_workspace}
\begin{figure}
    \begin{minipage}[c]{.46\linewidth}
        \centering
        \includegraphics[width=5cm]{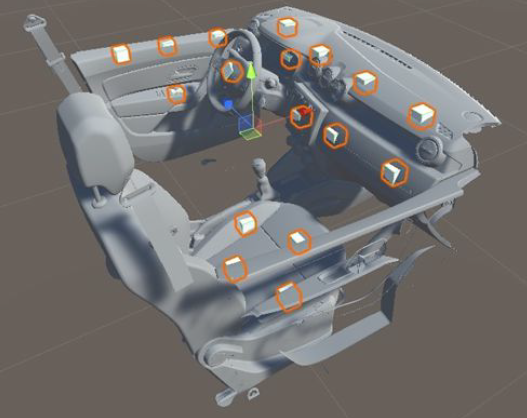}
        \caption{Unity VR system and representation of interaction points}
        \label{fig:VR_int_points}
    \end{minipage}
    \hfill%
    \begin{minipage}[c]{.46\linewidth}
        \centering
        \includegraphics[width=4cm]{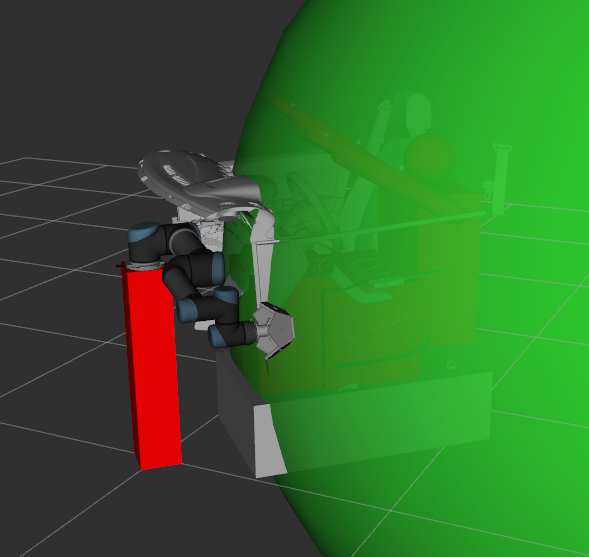}\\
        \caption{Representation of the user's effective workspace as a sphere}
        \label{fig:user_workspace_sphere}
    \end{minipage}
\end{figure}

%%%%%%%%%%%%%%%%%%%%%%%%%%
The first scenario is based on \cite{S_V_paper} where a distinction for velocity zones is made and where a plane divides the environment (space with the user and space where the user cannot go).% (Figure~\ref{fig:plane_workspace_plane}). 
Based on the same idea, we represented the effective working space of the mannequin as a sphere surrounding the model (Figure~\ref{fig:user_workspace_sphere}). 

The mobility scheme consists of alternating between different \emph{``Safe positions''}. These positions are so called because they are points out of reach of the user, which means that there is no need to constrain the robot's speeds. Therefore, the movement from one point to another just has to take into account the defined sphere, as we do not want to ``collide'' with it. Following this idea, we have performed a calculation of all existing trajectories between the different \emph{``Safe positions''} and stored them in a data file. This allows us to perform offline path planning, and then at runtime, depending on the initial and desired goal, we can access the pre-calculated paths to execute them directly, eliminating the computational time that would otherwise be required by performing online planning. The algorithm 1 allows the storage of the trajectory.

\begin{algorithm} [ht]
\label{alg:plan_storage_s1}
\caption{Trajectory computation and storage}
\begin{algorithmic} [1]
\REQUIRE{Number of points $no_p$. A counter for start point $i$. A counter for final point $j$}
\STATE $start\_name[no_p]$ $\leftarrow$ \COMMENT{Store id of the points}
\STATE $final\_name[no_p]$ $\leftarrow$ $init\_names[no_p]$ \COMMENT{Same id's as we are iterating through all points}
\FOR{ $i < 0$ ; $i < no_p$ ; $i++$ }
\FOR{ $j < 0$ ; $j < no_p$ ; $j++$ }
\IF{$i \neq j$}
\STATE $\O$ $\leftarrow$ Plan\_and\_Exec\_to($points[i]$) \COMMENT{Move to initial point of the plan}
\STATE $plan\_array[i][j]$ $\leftarrow$ Plan\_and\_Exec\_to($points[j]$) \COMMENT{Move to desired point and keep the planned trajectory}
\ENDIF
\ENDFOR
\ENDFOR
\STATE \COMMENT{Store all as a structured message}
\FOR{ $i < 0$ ; $i < no_p$ ; $i++$ }
\FOR{ $j < 0$ ; $j < no_p$ ; $j++$ }
\IF{$i \neq j$}
\STATE $init\_pos\_id$ $\leftarrow$ $start\_name[i]$
\STATE $goal\_pos\_id$ $\leftarrow$ $final\_name[j]$
\STATE $plan$ $\leftarrow$ $plan\_array[i][j]$
\ENDIF
\ENDFOR
\ENDFOR
\end{algorithmic}
\end{algorithm}

Then, the second part of the device consists of loading the pre-registered data and being able to use them on demand (Algorithm 2). We wait until we know the position we want to reach. Unlike  \cite{S_V_paper}, we have used a spherical surface here to divide the two areas of the space instead of a plane, as this allows greater flexibility for the planning group to consider more configurations when calculating the path between points. It also allows for more feasible trajectories for the robot.

\begin{figure}
%    \begin{minipage}[c]{.53\linewidth}
%        \centering
%        \includegraphics[width=4.5cm]{plane_object.png}
%        \caption{Representation of the user's effective workspace as a plane}
%        \label{fig:plane_workspace_plane}
%    \end{minipage}
%    \hfill%
%    \begin{minipage}[c]{.46\linewidth}
        \centering
        \includegraphics[width=3.5 cm]{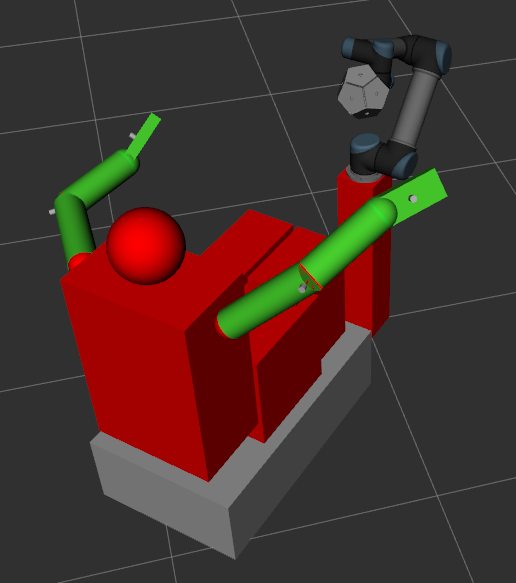}
        \caption{Moving obstacles}
        \label{fig:mov_obs}
%    \end{minipage}
\end{figure}

\begin{algorithm}[ht]
\label{alg:plan_exec_s1}
\caption{Trajectory upload and execution}
\begin{algorithmic} [1]
\REQUIRE{A desired frame to go to $des\_frame$. A home pose $home$. Number of elements $no_e$. Initial positions $init\_pos\_id$. Goal positions $goal\_pos\_id$. Planned trajectories $plan$. A counter $i$.}
\FOR{ $i < 0$ ; $i < no_e$ ; $i++$ }
\STATE \COMMENT{Extract the data from the file}
\STATE $start\_name[i]$ $\leftarrow$ $init\_pos\_id[i]$
\STATE $final\_name[i]$ $\leftarrow$ $init\_names[i]$ \COMMENT{Same id's as we are iterating through all points}
\STATE $plan\_array[i]$ $\leftarrow$ $plan[i]$ 
\ENDFOR
\STATE $\O$ $\leftarrow$ Plan\_and\_Exec\_to($home$) \COMMENT{Move to home pose}
\STATE \COMMENT{Reference to home position as current}
\STATE $init\_frame$ $\leftarrow$ $``home''$
\STATE $aux\_des\_frame$ $\leftarrow$ $``home''$
\WHILE{running}
    \IF{$des\_frame == init\_frame$}
        \STATE \COMMENT{The robot is in position.}

    \ELSE
        \STATE $aux\_des\_frame = des\_frame$ \COMMENT{Update the desired position}
        \FOR{ $i < 0$ ; $i < no_e$ ; $i++$ }
            \STATE \COMMENT{Search in the list of plans the one that matches the init and final frames}
            \IF{($start\_name[i]$ == $init\_frame$) \AND ($final\_name[i]$ == $aux\_des\_frame$)}
                \STATE $execute(plan\_array[i])$
                \STATE $init\_frame$ $\leftarrow$ $aux\_des\_frame$
            \ENDIF
        \ENDFOR
    \ENDIF
\ENDWHILE
\end{algorithmic}
\end{algorithm}
%%%%%%%%%%%%%%%%%%%%%%%%%
\subsubsection{Movement inside user's workspace}
%%%%%%%%%%%%%%%%%%%%%%%%%
The second mobility scheme has been proposed for the scenario where the robot end effector has to go inside the user's workspace, which means that the movements have to take into account the user's model in order to avoid any collision with him. We also have to take into account that the speed of these movements must be limited, in order to ensure safety.

Unlike the first scheme, in this case the environment consists of moving obstacles, which requires constant updating of the scene and constant tracking of the objects in it (Figure~\ref{fig:mov_obs}). For this reason, we used the images of the mannequin model to obtain its current positions and orientations in order to track their movement and link it to the objects created in the scene.

We also need to be able to determine whether a computed plan will collide or not, which requires taking several aspects into account. First, based on the calculated path to the desired goal, we check whether the path remains valid during the execution of the plan. To do so, we check for all calculated via-points of the path, whether the respective configurations are currently colliding with any other object present in the scene. If there are no collisions, we continue the execution. In the case of a collision present in any of the remaining states of the path, we instruct the robot to stop the execution of the computed path and replan it based on the updated scene information.

\begin{figure}
\centering
\includegraphics[width=3cm]{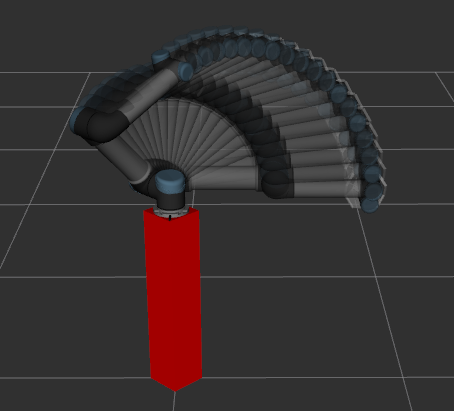}
\includegraphics[width=3cm]{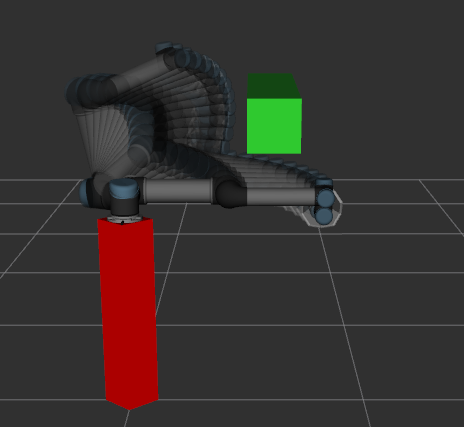}
\includegraphics[width=3cm]{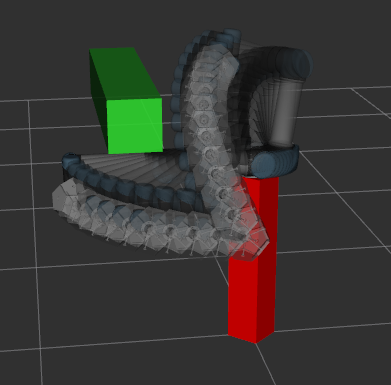}\\
\caption{ Original planned path (left) and Re-planned path from detected collision (middle and right)}
\label{fig:replan_path}
\end{figure}

To test our framework, we performed an initial trajectory planning. Then, during the execution, we created an obstacle. Then, by checking the validity of the trajectory, we are able to detect that an object is in collision with the planned trajectory. We then instruct the robot to stop the current execution and replan towards the same goal, taking into account the updated planning scene. This work is intended to be extrapolated to work according to the size of the mannequin. Thus, we can take into account the user moving in the environment as an obstacle to be avoided (Figure~\ref{fig:replan_path}).
%%%%%%%%%%%%%%%%%%%%%%%%%%%%%%%%%%%%%%%%%%%%%%%%%%%%%%%%%%%
\section{Conclusions}
%%%%%%%%%%%%%%%%%%%%%%%%%%%%%%%%%%%%%%%%%%%%%%%%%%%%%%%%%
In this paper, we have presented motion generation algorithms that can be used by a cobot to create an  intermittent contacts interface. A framework was presented including a UR5 cobot, ROS nodes, HTC Vive sensors and a car chair. Taking into account the objects present in the environment, a comparison of trajectory planning algorithms is presented. The selected algorithm is then used in two examples. An experimental validation is in progress and will be presented in the final version of the paper.
%%%%%%%%%%%%%%%%%%%%%%%%%%%%%%%%%%%%%%%%%%%%%%%%%%%%%%%%%%%
\section*{Acknowledgements}
%%%%%%%%%%%%%%%%%%%%%%%%%%%%%%%%%%%%%%%%%%%%%%%%%%%%%%%%%%%
This research is part of LobbyBot: Novel encountered type haptic devices, a French research project funded by ANR.
\bibliographystyle{IEEEtran}
\bibliography{HCI.bib}
\end{document}